\documentclass{article}

\usepackage[nonatbib, preprint]{neurips_2020}

\usepackage[utf8]{inputenc} 
\usepackage[T1]{fontenc}    
\usepackage{hyperref}       
\usepackage{url}            
\usepackage{booktabs}       
\usepackage{amsfonts}       
\usepackage{nicefrac}       
\usepackage{microtype}      
\usepackage{graphicx}
\usepackage{amsmath}
\usepackage{systeme}
\usepackage{amsthm}

\usepackage{graphicx}
\usepackage{mathrsfs}

\usepackage{xcolor}

\newtheorem{proposition}{Proposition}

\def\R{{\bf R}}
\def\N{{\bf N}}
\def\F{\mathscr{F}}
\def\S{\mathscr{S}}
\def\A{\mathscr{A}}

\DeclareMathOperator{\Th}{Th}

\def\x{\xi}
\def\s{\zeta}
\def\W{M}
\def\w{m}
\def\outputs{\eta}
\def\tvar{\tau}

\def\|#1|{\vbox{\hbox{\includegraphics{./const-#1.mps}}}}
\def\<#1>{\hbox{\includegraphics{./const-#1.mps}}}
\def\adj{\mathrel{\!\mathrel-\mkern-8mu\mathrel-\mkern-8mu\mathrel-\!}}
\newif\ifdraft

\ifdraft
\usepackage[notref,notcite]{showkeys}
\fi

\def\R{{\bf R}} 

\title{Backprop Diffusion is Biologically Plausible}

\author{%
Alessandro Betti$^{1}$, Marco Gori$^{1,2}$ \\
  $^{1}$DIISM, University of Siena, Siena, Italy \\
  $^{2}$Maasai, Universit\`{e} C\^{o}te d'Azur, Nice, France \\
  \texttt{\{alessandro.betti2,marco.gori\}@unisi.it} \\
}

\begin{document}

\maketitle

\begin{abstract}
The Backpropagation algorithm relies on the abstraction of using a neural
model that gets rid of the notion of time, since the input is mapped
instantaneously to the output.  In this paper, we claim that this abstraction
of ignoring time, along with the abrupt input changes that occur when feeding
the training set, are in fact the reasons why, in some papers, Backprop
biological plausibility is regarded as an arguable issue.  We show that as
soon as a deep feedforward network operates with neurons with time-delayed
response, the backprop weight update turns out to be the basic equation of a
biologically plausible diffusion process based on forward-backward waves. We
also show that such a process very well approximates the gradient for inputs
that are not too fast with respect to the depth of the network.  These
remarks somewhat disclose the diffusion process behind the backprop equation
and leads us to interpret the corresponding algorithm as a 
degeneration of a more general diffusion process that takes place also in
neural networks with cyclic connections.
\end{abstract}

\section{Introduction}
Backpropagation enjoys the property of being an optimal algorithm for
gradient computation, which takes $\Theta(m)$ in a feedforward network with
$m$ weights~\cite{Goodfellow:2016:DL:3086952,MachineLearningMG2018}.  It is
worth mentioning that the gradient computation with classic numerical
algorithms would take $O(m^2)$, which clearly shows the impressive advantage
that is gained for nowadays big networks. However, since its conception,
Backpropagation has been the target of criticisms concerning its biological
plausibility.  Stefan Grossberg early pointed out the {\em transport problem}
that is inherently connected with the algorithm.  Basically, for each neuron,
the delta error must be ``transported'' for updating the weights.  Hence, the
algorithm requires each neuron the availability of a precise knowledge of all
of its downstream synapses. Related comments were given by
F. Crick~\cite{Crick1989}, who also pointed out that backprop seems to
require rapid circulation of the delta error back along axons from the
synaptic outputs.  Interestingly enough, as discussed in the following, this
is consistent with the main result of this paper.
A number of studies have suggested solutions to the weight
transport problem.  Recently, Lillicrap et al~\cite{Lillicrap2016} have
suggested that random synaptic feedback weights can support error
backpropagation. However, any interpretation which neglects the role of time
might not fully capture the essence of biological plausibility.  The
intriguing marriage between energy-based models with object functions for
supervision that gives rise to Equilibrium
Propagation~\cite{DBLP:journals/neco/ScellierB19} is definitely better suited
to capture the role of time. Based the full trust on the role of temporal
evolution, in~\cite{DBLP:journals/tcs/BettiG16}, it is pointed out that,
like other laws of nature, learning can be formulated  under the framework of
variational principles.  

This paper springs out from recent studies
on the problem of learning visual features
\cite{tnnls2019,ijcai2019-278,DBLP:journals/corr/abs-1801-07110}
and it was also stimulated by a nice analysis on the interpretation of
Newtonian mechanics equations in the variational
framework~\cite{Stefanelli2013}.
%
It is shown that when shifting from algorithms to  
laws of learning, one can clearly see the emergence of the biological
plausibility of Backprop, an issue that has been controversial since its
spectacular impact. We claim that the algorithm does represent a sort of
degeneration of a natural spatiotemporal diffusion process that can
clearly be understood when thinking of perceptual tasks like speech and
vision, where signals possess smooth properties. In those tasks, instead of
performing the forward-backward scheme for any frame, one can properly spread
the weight update according to a diffusion scheme. While this is quite an
obvious remark on parallel computation, the disclosure of the degenerate
diffusion scheme behind Backprop, sheds light on its biological plausibility.
%
The learning process that
emerges in this framework is based on complex diffusion waves that, however,
is dramatically simplified under the feedforward assumption, where the
propagation is split into forward and backward waves.

\def\|#1|{\hbox{\includegraphics{./bio2-#1.mps}}}

\begin{figure}[t]
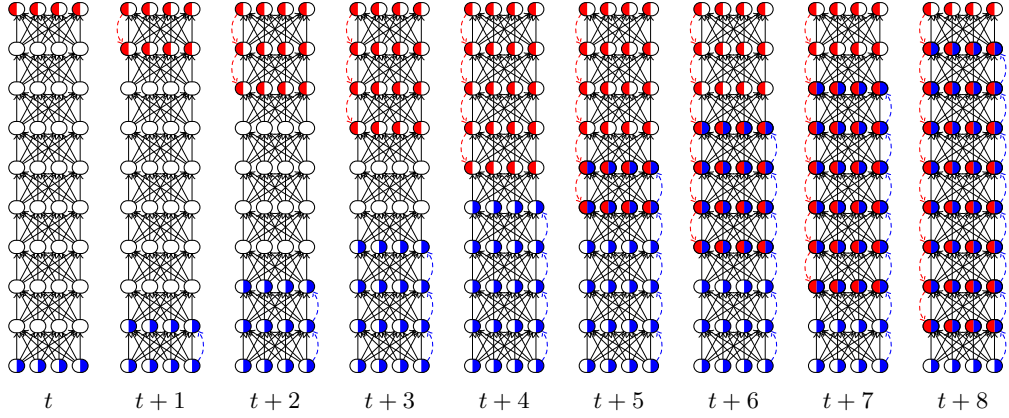

\footnotesize
\tabskip=1em plus2em minus.5em
\halign to \hsize{\hfil #\hfil& \hfil #\hfil&  \hfil #\hfil& 
\hfil #\hfil&  \hfil #\hfil &  \hfil #\hfil &  \hfil #\hfil &  \hfil #\hfil &  \hfil #\hfil \cr
\|1|&\|2|&\|3|&\|4|&\|5|&\|6|&\|7|&\|8|&\|9|\cr
\noalign{\smallskip}
$t$&$t+1$&$t+2$&$t+3$&$t+4$&$t+5$&$t+6$&$t+7$&$t+8$\cr}
\caption{\small Forward and backward waves on a ten-level network when the input and
the supervision are kept constant. When selecting a certain frame---defined
by the time index---the gradient can consistently be computed by
Backpropagation on ``red-blue neurons''.}
\label{fig1}
\end{figure}

\section{Backprop diffusion}
\label{BP-diff-Pic}
In this section we consider multilayered networks composed of $L$ layers of
neurons, but the results can easily be extended to any feedforward network
characterized by an acyclic path of interconnections.  The layers are denoted
by the index $l$, which ranges from $l=0$ (input layer) to $l=L$ (output
layer). Let $W_{l}$ be the layer matrix and $x_{t,l}$ be the vector of the
neural output at layer $l$ corresponding to discrete time $t$.
\begin{figure}[t!]
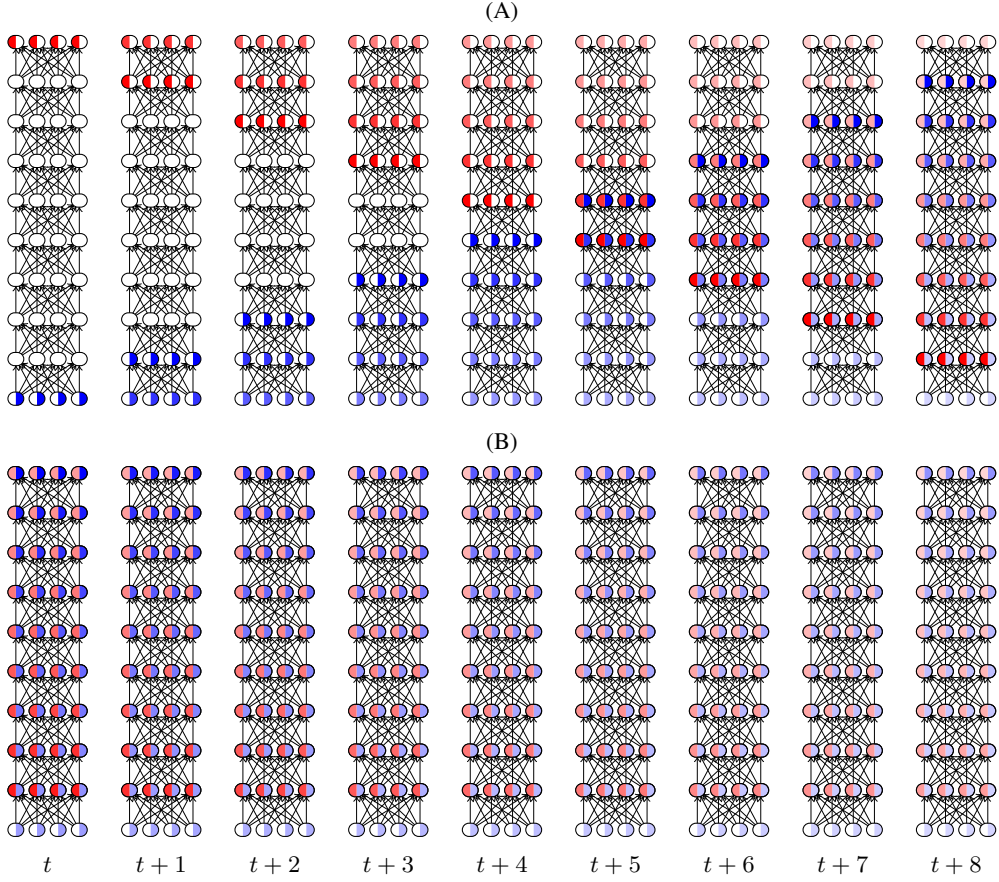

\footnotesize
\hbox to \hsize{\hfil(A)\hfil}
\smallskip
\tabskip=1em plus2em minus.5em
\halign to \hsize{\hfil #\hfil& \hfil #\hfil&  \hfil #\hfil& 
\hfil #\hfil&  \hfil #\hfil &  \hfil #\hfil &  \hfil #\hfil &  \hfil #\hfil &  \hfil #\hfil \cr
\|10|&\|11|&\|12|&\|13|&\|14|&\|15|&\|16|&\|17|&\|18|\cr
\noalign{\medskip}
\noalign{
\hbox to \hsize{\hfil(B)\hfil}
\smallskip}
\|28|&\|29|&\|30|&\|31|&\|32|&\|33|&\|34|&\|35|&\|36|\cr
\noalign{\smallskip}
$t$&$t+1$&$t+2$&$t+3$&$t+4$&$t+5$&$t+6$&$t+7$&$t+8$\cr}
\caption{\small Forward and backward waves on a ten-level network with slowly
varying input and supervision. In (A) it is shown how the input and
backward signals fill the neurons; the wavefronts of the waves are clearly
visible. Figure (B) instead shows the same diffusion process in stationary
conditions when the neurons are already filled up. In this quasi-stationary
condition the Backpropagation diffusion algorithm very well approximates the
gradient computation. In particular 
for $l_s=(L+1)/2=(9+1)/2=5$ there is a perfect backprop synchronization and the
gradient is correctly computed.}
\label{fig2}
\end{figure}
\begin{figure}[t]
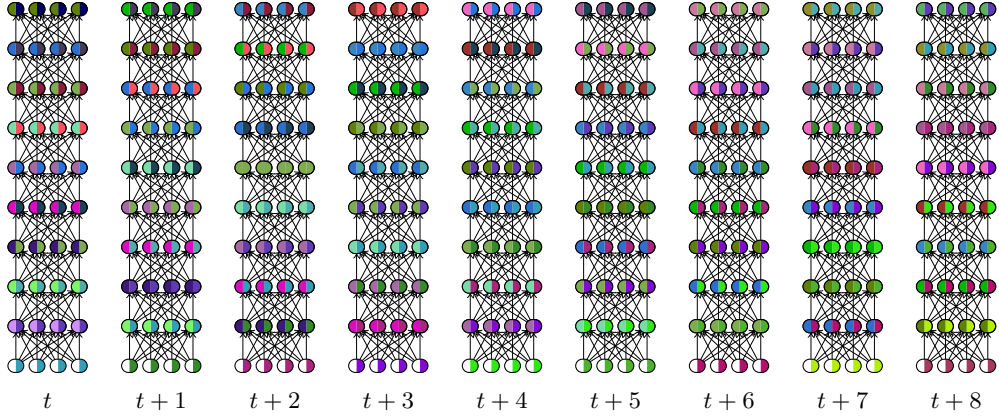

\footnotesize
\tabskip=1em plus2em minus.5em
\halign to \hsize{\hfil #\hfil& \hfil #\hfil&  \hfil #\hfil& 
\hfil #\hfil&  \hfil #\hfil &  \hfil #\hfil &  \hfil #\hfil &  \hfil #\hfil &  \hfil #\hfil \cr
\|19|&\|20|&\|21|&\|22|&\|23|&\|24|&\|25|&\|26|&\|27|\cr
\noalign{\smallskip}
$t$&$t+1$&$t+2$&$t+3$&$t+4$&$t+5$&$t+6$&$t+7$&$t+8$\cr}
\caption{Forward and backward waves on a ten-level network with rapidly changing
signal. In this case the Backpropagation diffusion equation does not properly
approximate the gradient. However, also in this case, for $l_s=5$ we have
perfect Backpropagation synchronization with correct syncronization of the
gradient.}
\label{fig3}
\end{figure}
Here we
assume that the network carries out a computation over time, so as, instead
of regarding the forward and backward steps as instantaneous processes, we
assume that the neuronal outputs follow the time-delay model:
\begin{equation}
x_{t+1,l+1} = \sigma(W_{l} x_{t,l}),    \label{OneStepDelayFS}
\end{equation}
where $\sigma(\cdot)$ is the neural non-linear function.  In doing so, when
focussing on frame $t$ the following forward process takes place in a deep
network of $L$ layers:
\begin{eqnarray}
\left\{
\begin{split}
	x_{t+1,1} &= \sigma(W_{0} x_{t,0}) \\
	x_{t+2,2} &= \sigma(W_{1} x_{t+1,1}) =  
	\sigma(W_{1}  \sigma(W_{0} x_{t,0})) \\
	&\,\,\,\vdots&\\
	x_{t+L,L} &= \sigma(W_{L-1} x_{t,L-1}) = \ldots 
	= \sigma(W_{L-1} \sigma(W_{L-2} \ldots \sigma(W_{0} x_{t,0}) \ldots)).
\end{split}\right.
\label{FS-Diffusion}
\end{eqnarray}
Hence, the input $x_{t,0}$ is forwarded to layers $1,2,\ldots,L$ a time
$t+1,t+2,\ldots,t+L$, respectively, which can be regarded as a forward wave.
We can formally state that input $u_{t}:=x_{t,0}$ is forwarded to layer
$\kappa$ by the operator $\stackrel{\kappa}{\rightarrow}$, that is
$\stackrel{\kappa}{\rightarrow}u_{t} = x_{t+\kappa,\kappa}$.
Likewise, when inspired by the backward step of Backpropagation, we can think
of back-propagating the delta error $\delta_{t,L}$ on the output as follows:
\[\left\{\begin{split}
\delta_{t+1,L-1} &= \sigma_{L-1}^{\prime} W_{L-1}^{T} \delta_{t,L} \\
\delta_{t+2,L-2} &= \sigma_{L-2}^{\prime} W_{L-2}^{T} \delta_{t+1,L-1}  
= (\sigma_{L-2}^{\prime} W_{L-2}^{T})
\cdot(\sigma_{L-1}^{\prime} W_{L-1}^{T}) \cdot \delta_{t,L}\\  
&\,\,\,\vdots&\\
\delta_{t+L-1,1} &=  \sigma_{1}^{\prime} W_{1}^{T} \delta_{t+L-2,2}
= \prod_{\kappa=1}^{L-1} \sigma_{\kappa}^{\prime} W_{\kappa}^{T} \cdot
                     \delta_{t,L}\end{split}\right.
\]
Like for $x_{t,l}$, we can formally state that the output delta error
$\delta_{t,L}$ is propagated back by the operator
$\stackrel{L-\kappa}{\longleftarrow}$ defined by
$\delta_{t,L} \stackrel{\kappa}{\longleftarrow} := 
\delta_{t+\kappa,L-\kappa}$.
The following equation is still formally coming from Backpropagation, since
it represents the classic factorization of forward and backward terms:
\begin{equation}
g_{t,l} = \delta_{t,l} \cdot x_{t,l-1}
= \big(\stackrel{l-1}{\longrightarrow} u_{t-l+1}\big)
\cdot \big(\delta_{t-L+l,L} \stackrel{L-l}{\longleftarrow}\big)
\label{DiffBP}
\end{equation}
%
%
Clearly, $g_{t,l}$ is the result of a diffusion process that
is characterized by the interaction of a forward and of a backward wave
(see Fig.~\ref{fig1}). 
This is a truly local spatiotemporal process which is definitely biologically plausible. Notice that if  $L$ is odd then for
$l_{s}=(L+1)/2$
we have a perfect 
backprop synchronization between the
input and the supervision, since in this case the number of
forward steps $l-1$ equals the number of backward steps $L-l$.
Clearly, the forward-backward wave synchronization takes place for 
$L=1$, which is a trivial case in which there is no wave propagation. 
The next case of perfect synchronization is for $L=3$. In this case,
the two hidden layers are involved in one-step of forward-backward 
propagation. Notice that the perfect synchronization comes with
one step delay in the gradient computation. In general, 
the computation of the gradient in the layer of perfect
synchronization is delayed of $l_{s}-1=(L-1)/2$. 
For all other layers, Eq.~\ref{DiffBP}
turns out to be an approximation of the gradient computation,
since the forward and backward waves 
meet in layers of no perfect synchronization. 
We can promptly see that, as a matter of fact, synchronization
approximatively holds whenever $u_{t}$ is not too fast with respect to
$L$ (see Fig.~\ref{fig2} and~\ref{fig3}).
The maximum mismatch between $l-1$ and $L-l$ is in fact $L-1$, so as if
$\Delta t$ is the quantization interval required to perform the computation
over a layer, good synchronization requires that $u_{t}$ is nearly constant
over intervals of length $\tau_{s}=(L-1)\Delta t$. For example, a video
stream, which is sampled at $f_{v}$ frames/sec requires to carry out the
computation with time intervals bounded by $\Delta t = \tau_{s}/(L-1) =
f_{v}/(L-1)$.

\section{Lagrangian interpretation of diffusion on graphs}
In the remainder of the paper we show that the forward/backward
diffusion of layered networks is just a special case of more
general diffusion processes that are at the basis of learning
in neural networks characterized by graphs with any pattern of
interconnections.
In particular we show that 
this naturally arises when formulating 
learning as a parsimonious constraint satisfaction problem.
We use recent connections established between learning processes and 
laws of physics under the principle of least cognitive action~\cite{DBLP:journals/tcs/BettiG16}.
\begin{table}[t]
\small
\begin{tabular}{ccl}  
\toprule
{\bf Learning}  & {\bf Mechanics} & {\bf Remarks} \\
\midrule
$(W,x)$       & $u$  & \vtop{\hsize 20pc
                       Weights and neuronal outputs
                       are interpreted as generalized
                       coordinates.}\\
\noalign{\smallskip}
$(\dot W,\dot x)$& $\dot u$ & \vtop{\hsize 20pc
                              Weight variations and neuronal variations
                              are interpreted as generalized velocities.}
                              \\
\noalign{\medskip}
  $\A(x,W)$ & $\S(u)$&  \vtop{\hsize 20pc
                       The cognitive action is the dual of the action in
                       mechanics.}
                       \\
\bottomrule
\end{tabular}
\medskip
\caption{Links between learning theory and classical mechanics.}
\label{table}
\end{table}

\indent
\parshape 7
0pc 28.5pc
0pc 27.5pc
0pc 27pc
0pc 27pc
0pc 27pc
0pc 27pc
0pc \hsize
\vadjust{\moveright 28.5pc\vbox to 0pt{\vskip -1pc
\hsize=0pt\vtop{\includegraphics{./ml-33.mps}}\vss}}
\noindent
In this paper we make the additional assumption of defining connectionist
models of learning in terms of a set of constraints that turn out to be
subsidiary conditions
of a variational problem~\cite{Giaquinta1996-1}.
Let us consider the classic example of the feedforward network used to
compute the
XOR predicate. If we denote with  $x^i$ the outputs of
each neuron and with $w_{ij}$ the weigh associated with the arch
$i\adj j$, then for the neural network in the figure we have
$x^3=\sigma(w_{31}x^1+w_{32}x^2)$, $x^4=\sigma(w_{41}x^1+w_{42}x^2)$
and $x^5=\sigma(w_{53}x^3+w_{54}x^4)$. Therefore in the
$w-x$ space this compositional relations between the nodes variables can be
regarded as constraints, namely $G^3=G^4=G^5=0$, where:
\[G^3=x^3-\sigma(w_{31}x^1+w_{32}x^2), \quad
G^4=x^4-\sigma(w_{41}x^1+w_{42}x^2),\]
\[G^5=x^5-\sigma(w_{53}x^3+w_{54}x^4).\]
In addition to these constraints we can also regard the way with which we
assign the input values as additional constraints. Suppose we want to
compute the value of the network on the input $x^1=e^1$ and $x^2=e^2$, where
$e^1$ and $e^2$ are two scalar values; this two assignments can be regarded as
two additional constraints $G^1=G^2=0$ where
$G^1=x^1-e^1,\quad G^2=x^2-e^2.$
First of all let us describe the architecture of the models that we will
address. Given a simple digraph $D=(V,A)$ of order $\nu$, without loss of
generality, we can assume $V=\{1,2,\dots,\nu\}$ and $A\subset V\times V$. A neural network constructed on $D$
consists of a set of maps
$i\in V\mapsto x^i\in \R$ and $(i,j)\in A\mapsto
w_{ij}\in \R$ together with $\nu$ constraints $G^j(x,W)=0$
$j=1,2,\dots \nu$ where $(W)_{ij}=w_{ij}$. Let ${\cal M}_\nu(\R)$ be the
set of all $\nu\times\nu$ real matrices and 
${\cal M}^{\downarrow}_\nu(\R)$ the set of all $\nu\times\nu$
strictly lower triangular matrices over $\R$. If
$W\in{\cal M}^{\downarrow}_\nu(\R)$ we say that the NN has a feedforward
structure. In this
paper we will consider both feedforward NN and NN with cycles. The relations
$G^j=0$ for $j=1,\dots, \nu$
 specify the computational scheme with which the information
diffuses trough the network. 
In a typical network with $\omega$ inputs these
constraints are defined as follows: 
For any vector $\x\in\R^\nu$, for any matrix $\W\in{\cal
M}_\nu(\R)$ with entries $\w_{ij}$ and for any given ${\cal C}^1$ map 
$e\colon (0,+\infty)\to\R^\omega$ we define the constraint on neuron $j$
when the example $e(\tau)$ is presented to the network as
\begin{equation}G^j(\tau,\x,\W):=\begin{cases}
\x^j-e^j(\tau), & \text{if $1\le j\le \omega$};\\
\x^j-\sigma(\w_{jk} \x^k) & \text{if $\omega <j\le \nu$},
\end{cases}\label{neuron-constraints-structure}\end{equation}
where $\sigma\colon \R\to\R$ is of class ${\cal C}^2(\R)$.
Notice that the dependence of the constraints on $\tau$ reflects
the fact that the computations of a neural network should be based on
external inputs.
%
%
%

\paragraph{Principle of Least Cognitive Action}
Like in the case of classical mechanics, when dealing with learning processes,
we are interested in the temporal dynamics of the variables 
exposed to the data from which the learning is supposed to happen.
%
Depending on the structure of the matrix $M$, it is useful to distinguish between feedforward networks and
networks with loops (recurrent neural networks). 
%
Let us therefore consider the functional
\begin{equation}
\A(x,W):=\int {1\over 2}(m_x\vert \dot x(t)\vert^2+m_W\vert \dot W(t)\vert^2)\,\varpi(t) dt+
\F(x,W),\label{cognitive-action}\end{equation}
with $\F(x,W):=\int F(t,x,\dot x,\ddot x,W,\dot W,\ddot W)\, dt$ and
$t\mapsto\varpi(t)$ a positive continuously differentiable function,
subject to the constraints
\begin{equation}G^j(t,x(t),W(t))=0,\qquad 1\le j\le \nu,
\label{neuron-constraints}\end{equation}
where the map $G(\cdot,\cdot,\cdot)$ is taken as in
Eq.~\eqref{neuron-constraints-structure}.
Let $({G_\x\atop \overline {G_\W}})$ be the Jacobian matrix of the constraints
$G$ with respect to $x$ and $W$, where it is intended that the first $\nu$
rows contain the gradients of $G$ with respect to its second argument:
\[\left({G_\x\atop \overline {G_\W}}\right)_{ij}\equiv G^j_{\xi^i},
\quad \hbox{for $1\le i\le\nu$}.
\]
Variational problems with subsidiary conditions can be tackled
using the method of Lagrange multipliers to convert the constrained problem into
an unconstrained one~\cite{Giaquinta1996-1}). In order to use this method it is
necessary to verify an independence hypothesis between the constraints;
in this case we need to check that the matrix $({G_\x\atop \overline {G_\W}})$
is full rank. Interestingly, the following proposition holds true:
\begin{proposition}
The matrix $({G_\x\atop \overline {G_\W}})\in {\cal M}_{(\nu^2+\nu)\times
\nu}(\R)$ is full rank.
\label{proposition}
\end{proposition}

\begin{proof}
First of all notice that if $(G_\x)_{ij}=G^j_{\x^i}$ is full rank
also $({G_\x\atop \overline {G_\W}})$ has this property. Then, since
\[G^j_{\x^i}(\tau,\x,\W)=\begin{cases}
\delta_{ij}, & \text{if $1\le j\le \omega$};\\
\delta_{ij}-\sigma'(\w_{jk} \x^k)\w_{ji} & \text{if $\omega <j\le \nu$,}
\end{cases}\]
we immediately notice that $G^i_{\x^i}=1$ and that for all $i>j$ we have
$G^i_{\x^i}=0$. This means that
\[
(G^j_{\x^i}(\tau,\x,\W))=\begin{pmatrix}
                          1&*&\cdots&*\\
                          0&1&\cdots&*\\
                          \vdots&\vdots&\ddots&\vdots\\
                          0&0&\cdots&1\end{pmatrix},\]
which is clearly full rank.
\end{proof}

Notice that this result heavily depends on the assumption
$W\in{\cal M}^{\downarrow}_\nu(\R)$ (triangular matrix), which
corresponds to feedforward architectures.
\paragraph{Derivation of the Lagrangian multipliers---feedforward networks}
Following the spirit of the principle of least cognitive action~\cite{DBLP:journals/tcs/BettiG16},
we begin by deriving
the constrained Euler-Lagrange (EL) equations associated with the
functional~\eqref{cognitive-action} under
subsidiary conditions~\eqref{neuron-constraints} that refer to feedforward neural networks.
The constrained functional is
\begin{equation}
\A\ ^*(x,W)=\int {1\over 2}(m_x\vert \dot x(t)\vert^2+m_W\vert \dot W(t)\vert^2)\varpi(t)
-\lambda_j(t)G^j(t,x(t),W(t))\, dt+\F(x,W),
\end{equation}
and its EL-equations thus read
\begin{align}
  &-m_x\varpi(t)\ddot x(t)-m_x\dot\varpi(t) \dot x(t)
  -\lambda_j(t)G^j_\x(x(t),W(t))
+L^x_F(x(t),W(t))=0;\label{x-eq}\\
  &-m_W\varpi(t)\ddot W(t)-m_W\dot\varpi(t) \dot W(t)
  -\lambda_j(t)G^j_\W(x(t),W(t))
+L^W_F(x(t),W(t))=0,\label{W-eq}
\end{align}
where $L_F^x=F_x-d(F_{\dot x})/dt+d^2(F_{\ddot x})/dt^2$,
$L_F^W=F_W-d(F_{\dot W})/dt+d^2(F_{\ddot W})/dt^2$
are the functional derivatives of $F$ with respect to $x$ and $W$
respectively (see \cite{courant-hilbert}). An expression for the Lagrange
multipliers is derived by differentiating two times the equations of the 
architectural constraints with respect to the
time and using the obtained expression to substitute the second order terms
in the Euler equations, so as we get:
\begin{equation}
\begin{aligned}
\Bigl({G^i_{\xi^a}G^j_{\xi^a}\over m_x}+{G^i_{m_{ab}}G^j_{m_{ab}}
\over m_W}\Bigr)\lambda_j=&
\varpi\bigl(G^i_{\tvar\tvar}+2(G^i_{\tvar \x^a}\dot x^a
+G^i_{\tvar \w_{ab}}\dot w_{ab}
+G^i_{\x^a\w_{bc}}\dot x^a\dot w_{bc})\\
&+G^i_{\x^a\x^b}\dot x^a \dot x^b
+G^i_{\w_{ab}\w_{cd}}\dot w_{ab}\dot w_{cd}\bigr)\\
&-\dot \varpi( \dot x^a G^i_{\xi^a}+\dot w_{ab} G^i_{m_{ab}})
+{L^{x^a}_FG^i_{\xi^a}\over m_x}+{L^{w_{ab}}_FG^i_{m_{ab}}\over m_W},
\end{aligned}\label{mult-lin-eq-x-w}
\end{equation}
where $G^i_\tvar$, $G^i_{\tvar\tvar}$, $G^i_{\xi^a}$, $G^i_{\xi^a\xi^b}$,
$G^i_{\w_{ab}}$ and  $G^i_{\w_{ab}\w_{cd}}$ are the gradients and the
hessians of constraint~\eqref{neuron-constraints}.

\medskip
\noindent
Suppose now that we want to solve Eq.~ \eqref{x-eq}--\eqref{W-eq}
with Cauchy initial conditions.
Of course we must choose $W(0)$ and $x(0)$ such that $g_i(0)\equiv 0$, where
we posed $g_i(t):= G^i(t, x(t), W(t))$, for $i=1,\dots,\nu$. However since the
constraint must hold also for all $t\ge0$ we must also have at least
$g'_i(0)=0$. These conditions written explicitly means
\[G^i_\tau(0,x(0),W(0))+G^i_{\xi^a}(0,x(0),W(0))\dot x^a(0)
+G^i_{m_{ab}}(0,x(0),W(0))\dot w_{ab}(0)=0.\]
If the constraints does not depend explicitly on time it is sufficient to
to choose $\dot x(0)=0$ and $\dot W(0)=0$, while for time dependent constraint
this condition leaves 
\[G^i_\tau(0,x(0),W(0))=0,\]
which is an additional constraint on the initial conditions $x(0)$ and $W(0)$
to be satisfied.
Therefore one possible consistent way to impose Cauchy conditions is
\begin{equation}
\begin{aligned}
&G^i(0,x(0),W(0))=0,\quad i=1,\dots,\nu;\\
&G^i_\tau(0,x(0),W(0))=0,\quad i=1,\dots,\nu;\\
&\dot x(0)=0;\\
&\dot W(0)=0.
\end{aligned}
\label{initial-cond}
\end{equation}

\paragraph{Reduction to Backpropagation}
To understand the behaviour of the Euler equations~\eqref{x-eq}
and~\eqref{W-eq} 
we observe that in the case of feedforward networks, as it
is well known, the constraints
$G^j(t,x,W)=0$ can be solved for $x$ so that eventually we can express the
value of the output neurons in terms of the value of the input neurons.
If we let $f^i_W(e(t))$ be the value of $x^{\nu-i}$ when $x^1=e^1(t),\dots
,x^\omega=e^\omega(t)$, then the theory defined by \eqref{cognitive-action} under
subsidiary conditions~\eqref{neuron-constraints} is equivalent, when
$m_x=0$ and $\varpi(t)=\exp(\vartheta t)$, to the 
unconstrained theory defined by
\begin{equation}
\int e^{\vartheta t}\Bigl(\frac{m_W}{2}\vert\dot W\vert^2-\overline V(t,W(t))
\Bigr)\, dt\label{supervised-unconstrained}
\end{equation}
where $\overline V$ is a loss function for which a possible choice is
$\overline V(t,W(t)):=\frac{1}{2}\sum_{i=1}^\outputs
(y^i(t)-f^i_W(E(t)))^2$ with $y(t)$ an assigned target.
The Euler equations associated
with~\eqref{supervised-unconstrained} are
\begin{equation}
\ddot W(t)+\vartheta \dot W(t)=-\frac{1}{m_W}\overline V_W(t,W(t)),
\label{euler-backprop}
\end{equation}
that in the limit $\vartheta\to\infty$ and $\vartheta m\to \gamma$ reduces to
the gradient method
\begin{equation}
\dot W(t)=-\frac1\gamma \overline V_W(t,W(t)),
\label{gradient-like-w}
\end{equation}
with learning rate $1/\gamma$.
Notice that the presence of the term $\varpi(t)$ that we proposed in the
general theory it is essential in order
to have a learning behaviour as it responsible of the
dissipative behaviour.

Typically the term $\overline V_W(t,W(t))$ in Eq.~\eqref{gradient-like-w}
can be evaluated using the Backpropagation algorithm; we will now show that
Eq.~\eqref{x-eq}--\eqref{mult-lin-eq-x-w} in the limit 
$m_x\to 0$, $m_W\to 0$, $m_x/m_W\to 0$ reproduces Eq.~\eqref{gradient-like-w}
where the term $\overline V_W(t,W(t))$ explicitly assumes the form
prescribed by BP. In order to see this choose $\vartheta=\gamma/m_W$,
$\varpi(t)=\exp(\vartheta t)$,
\[F(t,x(t),\dot x(t), \ddot x(t),W(t),\dot W(t),\ddot W(t))=-e^{\vartheta t}
V(t,x(t)),\]
and
multiply both sides of Eq.~\eqref{x-eq}--\eqref{mult-lin-eq-x-w} by
$\exp(-\vartheta t)$; then take the limit
$m_x\to 0$, $m_W\to 0$, $m_x/m_W\to 0$. In this limit Eq.~\eqref{W-eq} and
Eq.~\eqref{mult-lin-eq-x-w} becomes respectively
\begin{align}
&\dot W_{ij}=-\frac1\gamma  \sigma'(w_{ik}x^k) \delta_i x^j;\label{BP-grad}\\
&G^i_{\xi^a} G^j_{\xi^a} \delta_j=-V_{x^a}G^i_{\xi^a},\label{backward}
\end{align}
where $\delta_j$ is the limit of $\exp(-\vartheta t)\lambda_j$.
Because the matrix $G^i_{\x^a}$ is invertible Eq.~\eqref{backward} yields
\begin{equation}T\delta=-V_x,\label{backward-true}\end{equation}
where $T_{ij}:= G^j_{\xi^i}$. This matrix is an upper triangular matrix
thus showing
explicitly the backward structure of the propagation of the delta
error of the Backpropagation algorithm. Indeed in Eq.~\eqref{BP-grad}
the Lagrange multiplier $\delta$ plays the role of the delta error.

In order to better understand the perfect reduction of our approach to the backprop
algorithm consider the following example. We simply consider a feedforward network
with an input, an output and an hidden neuron. In this case the matrix $T$ is
\[
T=\begin{pmatrix}
1& -\sigma'(w_{21} x^1)w_{21}&0\\
0&1&-\sigma'(w_{32}x^2) w_{32}\\
0&0&1
\end{pmatrix}.
\]
Then according to Eq.~\eqref{backward-true} the Lagrange multipliers are
derived as follows
$\delta_3=-V_{x^3}, \delta_2=\sigma'(w_{32}x^2) w_{32}\delta_3$, and 
$\delta_1=\sigma'(w_{21}x^1) w_{21}\delta_2$.
This is exactly the Backpropagation formulas for the delta error. Notice that in
this theory we also have an expression for the multipliers of the input neurons,
even though, in this case, they are not used to update the weights
(see Eq.~\eqref{BP-grad}).

\paragraph{Diffusion on cyclic graphs}
The constraint-based analysis carried out so far assumes  holonomic constraints, 
whereas the claim of this paper is that we cannot neglect temporal dependencies,
which leads to expressing neural models by non-holonomic constraints.
In doing so, we go beyond  the constraints expressed by Eq.~\eqref{neuron-constraints}
which imply an infinite velocity of diffusion of information.
Since we are stressing the importance of time in learning processes, it
is natural to assume that the velocity
of information diffusion through a network is finite. In the discrete setting of computation
this is reflected by the model~\ref{OneStepDelayFS} discussed in Section~\ref{BP-diff-Pic}.
A simple classic translation of this constraint in continuous time is
\begin{equation}c^{-1}\dot x^i(t)=-x^i(t)+\sigma(w_{ik}(t) x^k(t)),
\label{first-nonhol}
\end{equation}
where $c>0$ can be interpreted as a ``diffusion speed''.
In stationary situation indeed Eq.~\eqref{first-nonhol} coincide with the
usual neuron equation in Eq.~\eqref{neuron-constraints}.
However, we are in front of a much more complicated
constraint, since not only it involves the variables $x$ and $w$, but also their 
derivatives. Such constraints are called non-holonomic constraints.
Now we snow show how to determine the stationarity conditions of the
functional\footnote{Notice that here
we are overloading the symbol $F$ since in this section
we assume that the Lagrangian $F$ depends only on $t$, $x$ and $W$.}
\[\A(x,W)=\int\left(\frac{m_x}{2}\vert\dot x(t)\vert^2+\frac{m_W}{2}
\vert \dot W(t)\vert^2+ F(t,x,W)\right)\, \varpi(t)\, dt\]
under the nonholomic constraints
\begin{equation}
G^i(t,x(t),W(t),\dot x(t)):=c^{-1}\dot x^i(t)+ x^i(t)
-\sigma(w_{ik}(t) x^k(t))=0,
\qquad i=1,\dots,r,
\end{equation}
by making use of the rule of the Lagrange multipliers. As usual we consider
the stationary points of the functional
\[\begin{split}
\A^*(x,W)=&\int\left(\frac{m_x}{2}\vert\dot x(t)\vert^2+\frac{m_W}{2}
\vert \dot W(t)\vert^2+ F(t,x,W)\right)\, \varpi(t)\, dt\\
&-\int \lambda_j(t) G^j(t,x(t),W(t),\dot x(t))\, dt.
\end{split}
\]
The Euler equations for $\A^*$ are
\[\begin{split}
&c^{-1}\dot x^i(t)+x^i(t)-\sigma(w_{ik}(t) x^k(t))=0;\\
&-m_W\varpi(t)\ddot W(t)-m_W\dot\varpi \dot W(t)
-\lambda_j G^j_M(t,x(t),W(t),\dot x(t))+\varpi(t) F_W(t,x(t),W(t))=0;\\
&-m_x\varpi(t)\ddot x(t)-m_x\dot\varpi \dot x(t)
+\dot\lambda_j G^j_p(t,x(t),W(t),\dot x(t))
+\lambda_j \frac{d}{dt}G^j_p(t,x(t),W(t),\dot x(t))\\
&\qquad-\lambda_j G^j_\xi(t,x(t),W(t),\dot x(t))+\varpi(t) F_x(t,x(t),W(t))=0.
\end{split}
\]
Again, if we assume that $F(t,x,W)=-V(t,x)$,
$\varpi(t)=e^{\vartheta t}$, $\delta_j=e^{-\vartheta t}\lambda_j$
and we explicitly use
the expression for $G^j_p$ we get
\[\begin{split}
&c^{-1}\dot x^i(t)+ x^i(t)-\sigma(w_{ik}(t) x^k(t))=0;\\
&-m_W\ddot W(t)-m_W\vartheta\dot W(t)
-\delta_j G^j_M(t,x(t),W(t),\dot x(t))=0;\\
&-m_x\ddot x(t)-m_x\vartheta\dot x(t)
+c^{-1}\dot\delta -\delta_j G^j_\xi(t,x(t),W(t),\dot x(t))-V_\xi(t,x(t))=0.
\end{split}
\]
This systems of equations can be better interpreted in the limit in which
we recovered the backprop rule in Section~4; that is to say in the
limit $m_x\to 0$, $m_x\to 0$ and $m_x/m_W\to 0$, $\theta\to\infty$ and
$\theta m_W=\gamma$ fixed. Under this conditions, we have the following further
reduction
\begin{equation}\label{c-diffusion}
\begin{split}
&c^{-1}\dot x^i(t)+x^i(t)-\sigma(w_{ik}(t) x^k(t))=0;\\
&\dot W(t)=-\frac{1}{\gamma}\delta_j(t) G^j_M(t,x(t),W(t),\dot x(t));\\
&c^{-1}\dot\delta(t)= \delta_j(t) G^j_\xi(t,x(t),W(t),\dot x(t))+V_\xi(t,x(t)).
\end{split}
\end{equation}
The equation that defines the Lagrange multipliers $\delta$
is now a differential equation that explicitly gives us the correct
updating rule instead of a instantaneous equation that must be solved
for each $t$. As the diffusion speed becomes infinite ($c\to+\infty$)
Eq.~\eqref{c-diffusion} reproduce Backpropagation, which
is consistent with the intuitive explanation that arises from 
Fig.~\ref{fig1},~\ref{fig2},~\ref{fig3}.
%


\section{Conclusions}
In this paper we have shown that the longstanding debate on the biological
plausibility of Backpropagation can simply be addressed by distinguishing the
forward-backward local diffusion process for weight updating with respect
to the algorithmic gradient
computation over all the net, which requires the transport
of the delta error through  the entire graph.
Basically, the algorithm expresses the
degeneration of a biologically plausible diffusion process, which comes from
the assumption of a static neural model. The main conclusion is that, more
than Backpropation, the appropriate target of the mentioned longstanding
biological plausibility issues is the assumption of an instantaneous map from
the input to the output. The forward-backward wave propagation behind
Backpropagation, which is in fact at the basis of the corresponding 
algorithm, is proven to be local and definitely biologically plausible. 
This paper has shown that an opportune embedding in time of deep networks
leads to a natural interpretation of Backprop as a diffusion process which is
fully local in space and time. The given analysis on any graph-based neural 
architectures suggests that spatiotemporal diffusion takes place according
to the interactions of forward and backward waves, which arise from the
environmental interaction. While the forward wave is generated by the 
input, the backward wave  arises from the output; the special way in 
which they interact for classic feedforward network corresponds to 
the degeneration of this diffusion process which takes place at infinite
velocity. However, apart from this theoretical limit, Backpropagation
diffusion is a truly local process.


\section*{Broader Impact}
Our work is a foundational study. We believe that there are neither
ethical aspects nor future societal consequences that should be discussed.

\bibliography{nn,corr}
\bibliographystyle{plain}

\end{document}